%% file: main.tex
\documentclass{article}

     \PassOptionsToPackage{numbers, compress}{natbib}

    \usepackage[preprint]{neurips_2020}

\usepackage[utf8]{inputenc} 
\usepackage[T1]{fontenc}    
\usepackage{hyperref}       
\usepackage{url}            
\usepackage{booktabs}       
\usepackage{amsfonts}       
\usepackage{amssymb}
\usepackage{nicefrac}       
\usepackage{microtype}      
\usepackage{comment} 
\usepackage{multirow}
\usepackage{colortbl}
\usepackage{paralist}
\usepackage{bbold}
\usepackage{graphicx}  
\usepackage[english]{babel}

\usepackage{amsmath}
\DeclareSymbolFontAlphabet{\amsmathbb}{AMSb}%
\newcommand{\R}[0]{\amsmathbb R}
\newcommand{\E}[0]{\amsmathbb E}

\DeclareMathOperator*{\relu}{ReLU}

\usepackage{xcolor}
\usepackage{algorithm}
\usepackage{algpseudocode}
\usepackage{bold-extra}
\usepackage{wrapfig}
\definecolor{deepcarminepink}{rgb}{0.94, 0.19, 0.22}
\definecolor{azure}{rgb}{0.0, 0.5, 1.0}

\title{Formatting Instructions For NeurIPS 2020}

\author{
   Yumo Xu\thanks{Work done during an internship at Microsoft.}\\
   School of Informatics\\
   University of Edinburgh, UK\\
   \texttt{yumo.xu@ed.ac.uk}\\
   \And
   Chenguang Zhu\\
   Microsoft Cognitive Services Research\\
   Redmond, WA, USA\\
   \texttt{chezhu@microsoft.com} 
   \And
   \hspace{0.5em}Baolin Peng\\
   \hspace{0.5em}Microsoft Research\\
   Redmond, WA, USA\\
   \hspace{0.5em}\texttt{bapeng@microsoft.com}
   \And
   \hspace{0.5em}Michael Zeng \\
   \hspace{0.5em}Microsoft Cognitive Services Research\\
   \hspace{0.5em}Redmond, WA, USA\\
   \hspace{0.5em}\texttt{nzeng@microsoft.com}
}

\begin{document}
\title{Meta Dialogue Policy Learning}
\maketitle
\begin{abstract}
Dialog policy determines the next-step actions for agents and hence is central to a dialogue system. However, when migrated to novel domains with little data, a policy model can fail to adapt due to insufficient interactions with the new environment. We propose Deep Transferable Q-Network (DTQN) to utilize shareable low-level signals between domains, such as dialogue acts and slots. We decompose the state and action representation space into feature subspaces corresponding to these low-level components to facilitate cross-domain knowledge transfer. Furthermore, we embed DTQN in a meta-learning framework and introduce Meta-DTQN with a dual-replay mechanism to enable effective off-policy training and adaptation.
In experiments, our model outperforms baseline models in terms of both success rate and dialogue efficiency on the multi-domain dialogue dataset MultiWOZ 2.0.
\end{abstract}

\section{Introduction}
\label{sec:intro}
Task-oriented dialogue systems aim to assist users to efficiently accomplish daily tasks such as booking a hotel or reserving dinner at a restaurant. Complex systems like Alexa and Siri often contain thousands of task domains. However, a successful model on one task often requires hundreds or thousands of carefully labelled domain-specific dialogue data, which consumes a large amount of human effort. Therefore, how to agilely adapt an existing dialogue system to new domains with a scant number of training samples is an essential task in task-oriented dialogues.

In this paper, we investigate dialogue policy, or dialogue management, which lies in the center of a task-oriented dialogue system. Dialogue policy determines the next-step action of the agent given dialogue states and the user's goals. As a dialogue is composed of multiple turns, the feedback to a dialogue policy's decision is often delayed until the end of the conversation. Therefore, Reinforcement Learning (RL) is usually leveraged to improve the efficiency and success rate in dialogue policy learning \citep{deepdynaq}.

There have been a number of methods applying dialogue policy in multi-domain settings \citep{peng2017composite,lipton2018bbq,lee2019convlab}. 
These models usually employ an all-in-one multi-hot representation for dialogue states. The state embedding vector is a concatenation of multiple segments, each as a multi-hot vector for the states in one domain. However, when there are unseen domains at inference time, 
the corresponding parameters of its dialogue acts and slots are not optimized. 
This significantly limits the adaptation performance of policy models.

To alleviate this problem, we note that there is often shareable low-level information between different domains. For instance, suppose the source domain is taxi-booking and the target domain is hotel-booking. Although the two domains have different ontologies, both domains share certain dialogue slots (e.g. \textit{start time} and \textit{location}) and dialogue acts (e.g. \textit{request} and \textit{inform}). These shared concepts bear a lot of similarities both in textual representation and corresponding agent policies. Thus, it is feasible to transfer domain knowledge via these commonalities in ontologies.

To this end, we propose a \textbf{D}eep \textbf{T}ransferable \textbf{Q}-\textbf{N}etwork (DTQN), based on Deep Q-Network (DQN) \citep{atari} in reinforcement learning, which learns to predict accurate Q-function values given dialogue states and system actions. In DTQN, we factorize the dialogue state space into a set of lower-level feature spaces. Specifically, we hierarchically model cross-domain relations at domain-level, act-level and slot-level. State representations are then composed of several shareable sub-embeddings. For instance, slots like \textit{start time} in different domains will now share the same slot-level embedding. Furthermore, instead of treating actions as independent regression classes as in DQN, we decompose the dialogue action space and our model learns to represent actions based on common knowledge between domains.

To adapt DTQN to few-shot learning scenarios, we leverage the meta-learning framework. Meta-learning aims to guide the model to rapidly learn knowledge from new environments with only a few labelled samples
\citep{maml,rakelly2019efficient}. 
Previously, meta-learning has been successfully employed in the Natural Language Generation (NLG) module in dialogues \citep{daml}. 
However, NLG is supervised learning by nature. 
Comparatively, there has been little work on applying meta-learning to the dialogue policy, as it is known that applying RL under meta-learning, a.k.a. meta-RL, is a much harder problem than meta supervised learning \citep{metarl}.

To verify this fact, we train the DTQN model under the Model-Agnostic Meta-Learning (MAML) framework \citep{maml}. However, we find through experiments that the canonical MAML fails to let the policy model converge because the task training phase leverages \textit{off-policy} learning while the task evaluation and meta-adaptation phase employ an \textit{on-policy} strategy. 
Thus, the model initially receives very sparse reward signals, especially on complex composite-domain tasks. As a result, the dialogue agent is prone to overfit the on-policy data and to get stuck at the local minimum in the policy space.

Therefore, we further propose Meta-DTQN with a \textit{dual-replay} mechanism. To support effective off-policy learning in meta dialogue policy optimization, we construct a task evaluation memory to cache dialogue trajectories and prefill it with rule-based experiences in task evaluation.
This dual-replay strategy ensures the consistency of off-policy strategy in both meta-training and meta-adaptation, and provides richer dialogue trajectory records to enhance the quality of the learned policy model. Empirical results show that the dual-replay mechanism can effectively increase the success rate of DTQN while reducing the dialogue length, and Meta-DTQN with dual replay outperforms strong baselines on the multi-domain task-oriented dialogue dataset MultiWOZ 2.0 \citep{budzianowski2018multiwoz}.



\section{Related Work}
\label{sec:rw}
\textbf{Dialogue Policy Learning}
Dialogue policy, also known as the dialogue manager, is the controlling module in task-oriented dialogue that determines the agent's next action. Early work on dialogue policy is constructed on manual rules \citep{ruledm}. As the outcome of a dialogue does not emerge until the end of the conversation, dialogue policy is often trained via Reinforcement Learning (RL) \citep{rlbook}. 
For instance, deep RL is proven useful for strategic  conversations \citep{drldm}, and
a sample-efficient online
RL algorithm is proposed to learn from only a few hundred dialogues \citep{sampleefficientdm}.
Towards more effective completion on complex tasks, hierarchical RL is employed to learn a multi-level policy either through temporal control \citep{peng2017composite}, or subgoal discovery \citep{tang2018subgoal}.
Model-based RL also helps a dialogue agent to plan for the future during conversations \citep{deepdynaq}. 
While RL for multi-domain dialogue policy learning has attracted increasing attention from researchers, dialogue policy transfer remains under-studied.

\textbf{Meta-Learning}
Meta-learning is a framework to adapt models to new tasks with a small number of data \citep{oneshotlearning}. It can be achieved either by finding an effective prior as initialization for new task learning \citep{oneshotlearning}, or by a meta-learner to optimize the model which can quickly adapt to new domains \citep{grant2018recasting}. Particularly, the model-agnostic meta-learning (MAML) \citep{maml} framework applies to any optimizable system. It associates the model's performance to its adaptability to new systems, so that the resulting model can achieve maximal improvement on new tasks after a small number of updates.

In dialogue systems, meta-learning has been applied to response generation. The domain adaptive dialog generation method (DAML) \citep{daml} is an end-to-end dialogue system that can adapt to new domains with a few training samples. It places the state encoder and response generator into the MAML framework to learn general features across multiple tasks.

\section{Problem Formulation}
\label{sec:problem}
\paragraph{Reinforced Dialogue Agent} Task-oriented dialogue management is usually formulated as a Markov Decision Process (MDP): a dialogue agent interacts with a user with sequential actions based on the observed dialogue states $s$ to fulfill the target conversational goal.
At step $t$, given the current state $s_t$ of the dialogue, the agent selects a system action $a_t$ based on its policy $\pi$, i.e., $a_t = \pi(s_t)$, and receives a reward $r_t$ from the environment\footnote{Reward $r$ measures the degree of success of a dialogue. In ConvLab \citep{lee2019convlab}, for example, success leads to a reward of $2*L$ where $L$ is the maximum number of turns in a dialogue (set to 40 in default), failure to a reward of $-L$. To encourage shorter dialogues, the agent also receives a reward of $-1$ at each turn.}.
The expected total reward of taking action $a$ under the state $s$ is defined as a function $Q(s,a)$:
\begin{equation}
    Q(s, a) = 
    \E_{\pi} 
    \left[
    \sum^{T-t}_{k=0}
    \gamma^k r_{t+k} \rvert
    s_t=s, a_t=a
    \right]
\end{equation}
where $T$ is the maximum number of turns in the dialogue, and $\gamma \in [0,1]$ is a discount factor.
The policy $\pi$ is trained to find the optimal Q-function $Q^{\ast}(s, a)$ so that the expected total reward at each state is maximized. The optimal policy is to greedily act as $\pi^{\ast} (s) = \text{argmax}_{a \in \mathcal{A}} Q^{\ast} (s, a)$.

To better explore the action space, an $\epsilon$-greedy policy is employed to select the action based on the state $s$: with probability $\epsilon$, a random action is chosen; with probability $1-\epsilon$, a greedy policy $a = \text{argmax}_{a'} Q^{\ast} (s, a'; \theta_Q)$ is taken.
Here, the Q-function is modeled by 
Deep Q-Network (DQN)
\citep{mnih2015human} with parameters $\theta_Q$. 
To train this network, state-action transitions $(s_t , a_t , r_t , s_{t+1})$ are stored in a \textit{replay buffer} $\mathcal{M}$. 
At each training step, a batch of samples is sampled from $\mathcal M$ to update the policy network via 1-step temporal difference (TD) error implemented with the mean-square error loss:
\begin{gather}
    \mathcal{L}(\theta_Q) = \E_{(s, a, r, s') \sim \mathcal{M}}
    \left[
    (y-Q(s, a;\theta_Q))^2 \right]
    \label{eq:loss}\\
    y = r + \gamma \text{max}_{a'}Q'(s', a'; \theta_{Q'})
\end{gather}
where $Q'$ is the \textit{target network} that is only periodically replaced by $Q$ to stabilize training.

\paragraph{Environment and Domain} 
The dialogue environment typically includes a database that can be queried by the system, and a \textit{user-simulator} that mimics human actions to interact with the agent.
At the beginning of a conversation,
the user-simulator specifies a dialogue \textit{goal}, and the agent is optimized to accomplish it. Dialogue goals are generated from one or multiple \textit{domain(s)}.
For instance, in the benchmark multi-domain dialogue dataset MultiWoz \citep{budzianowski2018multiwoz}, there are a total of 7 domains and 25 domain compositions. 
Each domain composition consists of one or more domains, e.g., \{\texttt{hotel}\} and \{\texttt{hotel}, \texttt{restaurant}, \texttt{taxi}\}. 
We split all domains into \textit{source} domains and \textit{target} domains to fit the meta-learning scenario (see Section \ref{sec:exp} for details).

\paragraph{State Representation}
We show the dialogue state representation for classic DQN in Figure \ref{fig:framework}(A).
After receiving a system action $a$, the environment 
responds
with a user action, which is then fed into a \textit{dialogue state tracker} (DST) to update the dialogue agenda. 
The DST maintains the entire dialogue records with a state dictionary,
and the DQN has a \textit{state encoder} to embed the dictionary into a state vector. 
In detail, this state encoder represents states with multi-hot state vectors including six primary feature categories \citep{lee2019convlab}, e.g.,  \textit{request} and \textit{inform}. 
As shown in the bottom-left corner of Figure \ref{fig:framework}(A), each category is encoded as the concatenation of a few domain-specific multi-hot vectors from its relevant domains, and the concatenation of the six category representations forms a binary state representation (see Appendix A for details).

We argue that two major issues in the classic DQN system prohibit its generalization to unseen domains: 
(1) the input states adopt multi-hot representations where no inter-state relation is considered and
(2) given the state input, actions in different domains are modeled as independent regression classes. 
However, there is a considerable amount of domain knowledge that can be shared across actions and states, e.g., both taxi-booking and hotel-reserving tasks share dialogue slots such as \textit{start time} and \textit{location} and dialogue acts such as \textit{request}. These types of information elicit similar text representation and policy handling.

\section{Framework}
\label{sec:framework}
\subsection{Deep Transferable Q-Network}
\label{subsection:dtqn}
\begin{figure*}[t]
  \centering
  \hspace*{-0.25in}
  \includegraphics[width=13.2cm]{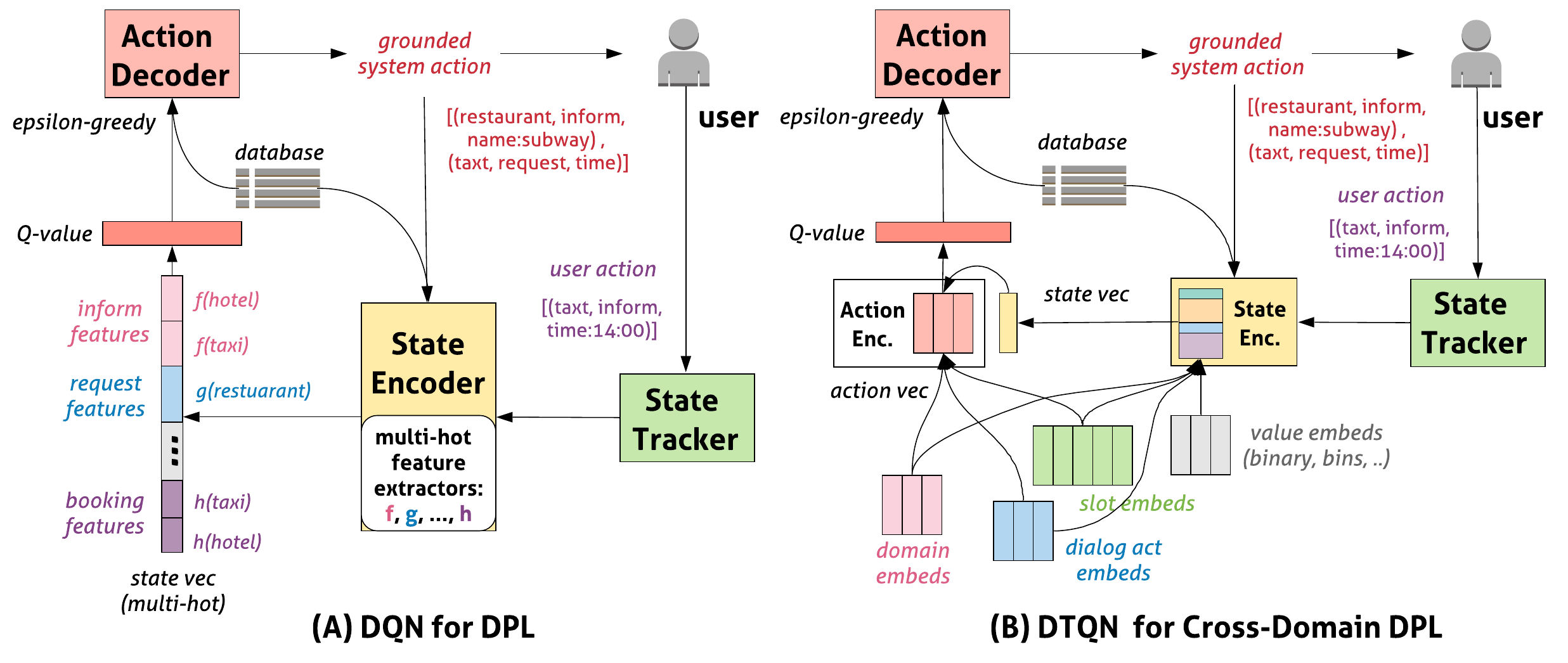}
  \caption{\label{fig:framework}Framework of (A) classic DQN for Dialogue Policy Learning and (B) our Deep Transferable Q-Network (DTQN) for Cross-Domain Dialogue Policy Learning.} 
\end{figure*}

To enable effective knowledge learning and transfer across different domains, we reformulate cross-domain dialogue policy learning as a state-action matching problem.
As shown in in Figure \ref{fig:framework}(B), we propose DTQN, a \textbf{D}eep \textbf{T}ransferable \textbf{Q}-\textbf{N}etwork that jointly optimizes the policy network and domain knowledge representations.

Driven by the structure of dialogue knowledge, we assume that the dialogue state space $\mathcal{S}$ and the system action space $\mathcal{A}$ are factorized by a set of lower-level feature spaces.
Based on this hypothesis, we aim at modeling cross-domain relations at different levels in DTQN: domain-level, act-level and slot-level. 
To this end, we hierarchically decompose the state and actions into four embedding subspaces, shared across all dialogue sessions: domains $\mathcal{D}$, dialogue acts $\mathcal{C}$, slots $\mathcal{O}$, and values $\mathcal{V}$. 
Both the states and actions are encoded by joining different sets of subspace embeddings.

We retain the existing categorization of dialogue state features mentioned in Section \ref{sec:problem} considering its effectiveness in dialogue management \citep{deepdynaq,lee2019convlab}.
We first represent each feature category as a dense vector $s_h, h\in [1,H]$ and then concatenate 
the $H=6$ categories of feature vectors
into the state representation $s\in \R^{d_s \times 1}$:
\begin{equation}
\label{eq:state}
    s = \relu\left(W_s[s_1, s_2, ..., s_H]\right).
\end{equation}
Note that each $s_h, h \in [1, H]$ consists of $|\mathcal{D}_h|$ domain-specific features, each corresponding to a domain. 
In the classic DQN state representation \citep{lee2019convlab}, few features are shared across domains.
As a result, an agent cannot generalize its policy from source domains to a target domain if the target state space remains unseen and the target action space is mostly unexplored.
Besides, the length of state representations grows linearly with $|\mathcal{D}_h|$.

Here, we propose to use a fixed-length state vector $s_h$ to represent the $h$-th feature category. To do that, we use cross-domain features to aggregate state information from different domains. 
In detail, the $i$-th domain-specific component of $s_h$ is denoted by $\hat{s}_{h,i}$. 
For example, for the \textit{inform} category,
\begin{equation}
    \hat{s}_{h,i} = \left[d_{h,i}, 
    \overline{o \odot v},
    u_{h,i}
    \right]
\end{equation}
where $d_{h,i}$ denotes embedding of domain. 
$\overline{o \odot v}$ is the average of the inner product between general slot embeddings and their value embeddings.
The binary feature $u_{h,i}$ tracks whether 
the corresponding domain
is active, i.e.,  essential domain slots are already filled.

To obtain the fixed-length representation for a feature category, we aggregate its domain features from all relevant domains via a non-linear transformation with a residual connection: 
\begin{equation}
    s_h = \frac{1}{|\mathcal{D}_h|}\sum_{i=1}^{|\mathcal{D}_h|} 
\left(
\hat{s}_{h,i} + \text{ReLU}
\left(W_h \hat{s}_{h,i}+b_h\right)
\right)
\end{equation}
where $W_h$ projects domain-specific features into a shared feature space across domains and we acquire the final state representation $s$ via Equation~(\ref{eq:state}).

Different from DQN, which encodes only dialogue states and incorporates no prior information of actions, we explicitly model the structural information of system actions with an \textit{action encoder} in DTQN to maximize knowledge sharing across domains.
Action encoding follows analogous procedures to the state encoding except that it does not use value space $\mathcal V$.
For each system action $a$, the domains that contain this action form a set $\mathcal{D}_a$. We encode its $\ell$-th domain feature ($1\leq {\ell} \leq |\mathcal{D}_a|]$) as
$\hat{a}_{\ell} = [d_{\ell}, c_{\ell}, \overline{o}]$,
where $c_{\ell}$ is embedding for dialogue act, e.g., 
\textit{request} or \textit{booking}, and $\overline{o}$ is the average of slot embeddings.
We then obtain the system action embedding: $a = 1/|\mathcal{D}_a|\sum_{\ell=1}^{|\mathcal{D}_a|}
\left(
\hat{a}_{\ell} + \text{ReLU}\left(W_a \hat{a}_{\ell}+b_a\right)
\right)$.


All embedding tables are shared between state and action encoders. 
We stack all action vectors and denote the action matrix as $A\in \R^{|\mathcal{A}|\times d_a}$, which is then used to produce the Q-values:
\begin{equation}
    Q(s, a) = \frac{1}{\sqrt{d_a}}A W_q s \in \R^{|\mathcal{A}|}
\end{equation}
where $W_q \in \amsmathbb{R}^{d_a \times d_s}$ is a parameter matrix.


\subsection{Meta Reinforcement Learning with Dual Replay}
\input{algo_core}
To adapt Q-network to few-shot learning scenarios, we propose to use a meta-learning framework \citep{maml} and present an instantiation of this framework with the DTQN as the policy network $Q$ and target network $Q'$.
Algorithm~\ref{algo} shows the pseudocode for our methodology for meta dialogue policy learning. 

At the beginning of each outer-loop of meta-training, we first sample $K$ dialogue goals as training tasks. For the $k$th inner-loop step,
the agent interacts with the environment using task $t_k$ to collect trajectories and stores them in the replay buffer $\mathcal{M}_{tr}$ (see Appendix B for details of function \textsc{EnvInteract}). Then, we sample from $\mathcal{M}_{tr}$ a minibatch of experiences of task $t_k$: $\mathcal{B}_{tr}^{t_k}$. The loss function $\mathcal{L}_{t_k}$ is from Equation ~(\ref{eq:loss}).
We compute task-specific updated parameters $\theta_{Q}^{(k)}$ from $\theta_{Q}$:
\begin{gather}
\theta_{Q}^{(k)}
=\theta_{Q}-
\alpha \nabla_{\theta_{Q}} \mathcal{L}_{t_{k}}\left(\theta_{Q}; \mathcal{B}_{tr}^{t_k} \sim \mathcal{M}_{tr}\right) \text{ where }\\
\nabla_{\theta_{Q}} \mathcal{L}_{t_k}\left(\theta_{Q}\right)=\E_{\left(s, a, r, s^{\prime}\right) \sim \mathcal{B}_{tr}^{t_k}}\left[
(r+ 
\gamma \max _{a^{\prime}}
Q^{\prime}\left(s^{\prime}, a^{\prime} ; \theta_{Q^{\prime}}\right)
-
Q\left(s, a ; \theta_{Q}\right)
) 
\nabla_{\theta_{Q}} 
Q\left(s, a ; \theta_{Q}\right)
\right].
\end{gather}
With the updated parameters $\theta_{Q}^{(k)}$, the agent interacts with the environment and obtains trajectory $\mathcal{B}_{ev}^{t_k}$.
According to MAML \citep{maml}, the task evaluation loss $\mathcal{L}_{t_{k}}(\mathcal\theta_{Q}^{(k)};
   \mathcal{B}_{ev}^{t_k})$ should be directly used to update $\theta_Q$ with learning rate $\beta$:
\begin{equation}
\label{eq:eval}
   \theta_{Q} \leftarrow \theta_{Q} -\beta \nabla_{\theta_{Q}} \sum_{k=1}^K \mathcal{L}_{t_{k}}(\mathcal\theta_{Q}^{(k)};
   \mathcal{B}_{ev}^{t_k}).
\end{equation}
However, this \textit{on-policy} learning suffers from very sparse rewards especially at the initial learning stage. 
This is due to the inherent difficulties in cross-domain dialogue learning: i) the state-action space to explore is much larger, and ii) the conversation required to complete the task is often longer \citep{peng2017composite}.
As a result, the dialogue agent is prone to overfit with on-policy data and to get stuck at the local minimum in the policy space. 


To alleviate this problem, we propose a dual-replay framework to support efficient \textit{off-policy} learning in meta-RL.
Apart from the main replay buffer $\mathcal{M}_{tr}$ for meta-training, we construct a task evaluation memory $\mathcal{M}_{ev}$. We note that it is essential to separate $\mathcal{M}_{tr}$ and $\mathcal{M}_{ev}$ since the task replay buffer is for the evaluation purpose for each task and should not be seen during task training.

Moreover, we adopt a variant of imitation learning, Replay Buffer Spiking (RBS) \citep{lipton2016efficient} to warm up the learning process.
Before our agent interacts with the environment, we employ a rule-based agent crafted for MultiWoz to initialize both $\mathcal{M}_{tr}$ and  $\mathcal{M}_{ev}$.
Then, in steps \ref{algo:sample}-\ref{algo:loss}, we collect new trajectories with our agent and push them into $\mathcal{M}_{ev}$. 
We uniformly sample from $\mathcal{M}_{ev}$ a mini-batch $\mathcal{B}_{ev}^{t_k}$, which can be a mixture of on-policy and relevant off-policy data, to calculate the task evaluation loss $\mathcal{L}_{t_k}$. As a result, $\theta_Q$ is updated as:
\begin{equation}
   \theta_{Q} \leftarrow \theta_{Q} -\beta \nabla_{\theta_{Q}} \sum_{k=1}^K \mathcal{L}_{t_{k}}(\mathcal\theta_{Q}^{(k)}; \mathcal{B}_{ev}^{t_k} \sim \mathcal{M}_{ev}).
\end{equation}
During test, for an unseen domain, we adopt a similar off-policy approach for meta-adaptation. This train-test consistency circumvents the known difficulty in on-policy meta-adaptation with off-policy meta-training \citep{rakelly2019efficient}.
In fact, classic MAML for RL can be seen as a special case of our dual-replay architecture by setting the task evaluation memory $\mathcal{M}_{ev}$ to $|\mathcal{B}_{ev}^{t_k}|$. 

\section{Experiment}
\label{sec:exp}
\subsection{Setup}
\input{table_tgt}
\paragraph{Dataset and task settings}
We use the benchmark multi-domain dialogue dataset MultiWoz 2.0 \citep{budzianowski2018multiwoz} for the evaluation. We adopt \texttt{attraction}, \texttt{restaurant}, \texttt{taxi} and \texttt{hospital} as source domains for training (source task size $K=4$), and use \texttt{hotel}, \texttt{train} and \texttt{police} as target domains for adaptation.
This split makes sure that both train and test splits have domains with various frequency levels (see Appendix C for details).
We propose two experiment settings: \textit{single-domain} and \textit{composite-domain}.
For the \textit{single-domain} setting, agents are trained and tested with only single-domain dialogue goals. 
In the \textit{composite-domain} setting, for each task in meta-training, we first select a seed domain $d^{\ast}$ and then sample domain composition which contains $d^{\ast}$. The trained model is then adapted and evaluated in various domain compositions containing $d^{\ast}$. 

\paragraph{Systems}
We developed different baseline task-oriented dialogue systems: \textsc{Dqn} is standard deep Q-learning which uses binary state representations and \textsc{Dtqn}, which is our proposed model without meta-learning framework.
We also build \textsc{VanillaDqn} without Replay Buffer Spiking (RBS) \citep{lipton2016efficient}
to show the warm-up effects in adaptation from rule-based off-policy data.
In addition, we build \textsc{Meta-Dtqn-Sr} with only one single replay buffer to show the effects of the dual-replay mechanism we proposed.
During adaptation to target domains, we simulate the data scarcity scenario by using only 1,000 frames (i.e., 10\% of the training data).
Besides, to examine the effects of the two-stage paradigm of training-and-adaptation, we also report the results of two few-shot models, \textsc{Dqn-1k} and \textsc{Dtqn-1k}. Both models are trained from scratch with the 1,000 frames in the target domains.

\paragraph{Implementation Details} 
We developed all variants of agents based on Convlab \citep{lee2019convlab}.
We used a batch size of 16 for both training and adaptation. 
We set the size of the training replay buffer and the evaluation replay buffer to 50,000. 
We initialized the replay buffers with Replay Buffer Spiking (RBS) \citep{lipton2016efficient} during the first 1000 episodes of meta-training, first 10 episodes of single-domain adaptation and first 50 episodes of composite-domain adaptation (see Appendix D for details).

\subsection{Evaluation Results}
\input{table_src}
\input{fig_test_replay}
Table \ref{tab:tgt_single} shows that our models (\textsc{Meta-Dtqn} and \textsc{Dtqn}) considerably outperform baseline systems on single-domain tasks in \texttt{hotel} and \texttt{train}. 
Dialogue tasks in \texttt{police} are relatively easy to accomplish, on which \textsc{DQN-1k} trained from scratch with only 1,000 frames can completely succeed. On the contrary, \textsc{DQN-1k} fails on all the tasks in \texttt{Hotel}.
Also, note that \textsc{Dtqn-1k} significantly outperforms \textsc{Dqn-1k} across all domains. This demonstrates the effectiveness of modeling the dependency between the state and action space. 
Besides, the performance gain from meta-training is more significant in the \texttt{train} domain (i.e., 4.8\% in success rate), which can be attributed to the similarity of state and action spaces between \texttt{train} and the source domain \texttt{taxi}.

Table \ref{tab:tgt_cross} shows the adaptation results on the composite-domain setting, which is a much harder dialogue task. Here, \textsc{Meta-Dtqn} has a clear advantage over other agents on both \texttt{hotel} and \texttt{train}, showing that meta-learning can boost the quality of agents adapted with a small amount of data in complex dialogue tasks (see Appendix~E for dialogue examples).
Table \ref{tab:src} lists the performance of various models when evaluating on source domains. Here, meta-learning can also help to achieve better results, and the gain is larger in the more complex composite-domain settings.

It is worth noting that on all tasks,
\textsc{Meta-Dtqn} shows superior results than its single-replay counterpart, \textsc{Meta-Dtqn-Sr}, and the performance gap is particularly large on   composite-domain dialogue tasks where the agent is more prone to suffer from initial reward sparsity.

\textbf{Effects of dual replay} We further investigate the effects of the proposed dual-replay method.
In Figure \ref{fig:test_mem_size}, we show the performance of our model with a task evaluation memory of varied sizes. 
We start with pure on-policy evaluation
$\left|\mathcal{M}_{ev}\right|=16$, i.e., batch size $|\mathcal{B}_{ev}^{t_k}|$, 
and experiment with different buffer sizes: 16, 1000, 3000, 5000, 10000, and 50000.
As shown, when the replay buffer is relatively small ($<5000$), the success rate fails to improve.
We argue that this optimization difficulty is due to the overfitting to on-policy data with sparse rewards at the beginning of the learning phase. This can be verified by the loss curve: the training loss abruptly drops from high values (100-500) to extremely low values (less than 10) soon after the RBS warm-up phase.
When the evaluation memory size increases, our model is able to escape from the local minimum and get optimized continuously.

\input{fig_ada_data}
\textbf{Effects of adaptation data size} In addition, we show how the size of adaptation data affects the agent's performance on target domains in Figure \ref{fig:ada_data}. We test \textsc{Meta-Dtqn} with adaptation data ranging from 100 frames (1\% of the training data) to 2,500 frames (25\% of the training data). 
As shown, the agent performance positively correlates with the amount of data available in the target domain. Note that as one episode has on average 10 frames, and we adopt RBS for the first 50 episodes, the agent is adapted only with off-policy rule-based experiences when the number of frames is less than 500. Therefore, the large performance gap between 500 and 1,000 frames indicates that our model can considerably benefit from a very small amount of on-policy data. 


\section{Conclusion}
\label{sec:conclusion}
Dialogue policy is the central controller of a dialogue system and is usually optimized via reinforcement learning. However, it often suffers from insufficient training data, especially in multi-domain scenarios. In this paper, we propose the Deep Transferable Q-Network (DTQN) to share multi-level information between domains such as slots and acts. We also modify the meta-learning framework MAML and introduce a dual-replay mechanism. Empirical results show that our method outperforms traditional deep reinforcement learning models without domain knowledge sharing, in terms of both success rate and length of dialogue. As future work, we plan to generalize our method to more meta-RL applications in multi-domain and few-shot learning scenarios.

\section*{Broader Impact}
Our work can contribute to dialogue research and applications, especially in new domains with scant training data. Our framework helps models quickly adapt to unseen domains to bootstrap applications. The outcome is a more effective and efficient dialogue agent system to facilitate activities in human society.

However, one needs to be cautious when collecting dialogue data, which may cause privacy issues. Deanonymization methods must be used to protect personal privacy.

\bibliographystyle{unsrt}

\end{document}

%% file: algo_core.tex
\begin{algorithm}[t]
\small
\caption{\label{algo}Meta Dialogue Policy Learning}%
\begin{algorithmic}[1]
\Function{MetaPolicyLearning}{}
\State Initialize $Q(s, a; \theta_Q)$ and $Q'(s, a; \theta_{Q'})$ with $\theta_{Q'}\gets\theta_{Q}$  \Comment{Policy network and target network}
\State Initialize experience replay memory $\mathcal M_{\textit{tr}}$ and $\mathcal M_{\textit{ev}}$ using Reply Buffer Spiking (RBS)  \Comment{Dual replay}
\State Set $K$ domains $\{d_k\}^K_{k=1}$ and gather the domain compositions $\{\mathcal{T}_{d_k}\}^K_{k=1}$, where $d_k \in \mathcal{T}_{d_k}, 1\leq k \leq K$.

\For{$n\gets 1:N$} \Comment{Outer loop for meta-training}
    \State Generate $K$ dialogue goals $ \{t_1, ..., t_K\}$ from $d_k$ or $\mbox{Uniform} (\mathcal{T}_{d_k})$
    \Comment{Single or composite domain}
    \State Initialize meta-training loss $\mathcal{L} \gets 0$ 
    \For{$k\gets1:K$} \Comment{Inner loop for task data collection and training}
        \State $\theta' \gets \theta$ and load agent with $\theta'$
        \State \Call{EnvInteract}{$t_{k}$, $\mathcal M_{\textit{tr}}$, $B_{\textit{tr}}$}
        \Comment{Task training data collection}
        \State Sample random minibatches of $(t_k, s, a, r, s')$ from $\mathcal M_{\textit{tr}}$
        \State Update $\theta'$ via $Z$-step minibatch SGD
        \State \Call{EnvInteract}{$t_{k}$, $\mathcal M_{\textit{ev}}$, $B_{\textit{ev}}$}
        \Comment{Task evaluation data collection}
        \State Sample random minibatches of $(t_k, s, a, r, s')$ from $\mathcal M_{ev}$\label{algo:sample}
        \State Forward pass with the minibatches and obtain $\mathcal{L}_{t_k}$
        \State $\mathcal{L} \gets \mathcal{L}+\mathcal{L}_{t_k}$\label{algo:loss}
    \EndFor
    \State Load agent with $\theta_Q$ and update with respect to $\mathcal{L}$ via minibatch SGD
    \State Every $C$ steps reset $\theta_{Q'}\gets \theta_{Q}$ \Comment{Target network update}
\EndFor
\EndFunction
\end{algorithmic}
\end{algorithm}

%% file: table_tgt.tex
\begin{table}[t]
\tabcolsep=0.03cm
\small
\bgroup
\def\arraystretch{1.0}
\centerline{
\begin{tabular}{l|ccc|ccc|ccc|ccc}
\hline
\multirow{2}*{\textbf{Systems}} & \multicolumn{3}{c}{\texttt{Hotel}} & \multicolumn{3}{c}{\texttt{Train}} & \multicolumn{3}{c}{\texttt{Police}} &
\multicolumn{3}{c}{Average}\\
~ & 
\textbf{Success} & \textbf{Reward} & \textbf{Turns} & 
\textbf{Success} & \textbf{Reward} & \textbf{Turns}& 
\textbf{Success} & \textbf{Reward} & \textbf{Turns}& 
\textbf{Success} & \textbf{Reward} & \textbf{Turns}\\
\hline
\rowcolor[gray]{0.95}\multicolumn{13}{c}{\textbf{Few-Shot Models}}\\
\hline
\textsc{Dqn-1k} 
&\hspace{1.5ex}0.00	&\hspace{-0.5ex}-55.70	&17.71	&\hspace{1ex}2.15	&\hspace{-0.5ex}-56.02	&20.60	&\textbf{100.00}	&76.62	&5.38
&14.66	&\hspace{1ex}8.58	&13.68\\
\textsc{Dtqn-1k}
&53.90	&15.06	&11.62	&61.00	&24.84	&10.36	&\textbf{100.00}	&\textbf{79.28}	&\textbf{2.72}
&71.63	&39.73	&\hspace{1ex}8.23
\\
\hline
\rowcolor[gray]{0.95}\multicolumn{13}{c}{\textbf{Adaptive Models}}\\
\hline
\textsc{VanillaDqn}
&28.10	&\hspace{-0.5ex}-20.18	&15.89	
&\hspace{1ex}0.00	&\hspace{-0.5ex}-59.00	&21.00	
&\hspace{1ex}24.30	&\hspace{-0.75ex}-25.95	&\hspace{-1.25ex}17.11
&17.47	&\hspace{-0.75ex}-35.04	&18.00\\
\textsc{Dqn} 
&36.00	&\hspace{0.5ex}-9.70   &14.90	&32.20	&\hspace{-0.5ex}-14.53   &15.17	&\textbf{100.00} &76.62  &5.38
&56.07	&17.46	&11.82
\\
\textsc{Dtqn}
&\textbf{62.70}   
&\textbf{27.99}  &\hspace{1ex}\textbf{9.25}
&82.65  
&54.80   
&\hspace{1ex}6.38
&\textbf{100.00}
&{79.24}
&{2.76}
&81.78	&54.01&	\hspace{1ex}6.13\\
\textsc{Meta-Dtqn-Sr}
&47.50&	\hspace{1ex}5.86&	13.14
&50.35&	10.26	&12.16
&\textbf{100.00}&	\textbf{79.28}	&\textbf{2.72}
&65.95&	31.80	&\hspace{1ex}9.34\\
\textsc{Meta-Dtqn}
&{61.90}
&{26.28}
&{10.00}
&\textbf{87.45}
& \textbf{61.44}   &\hspace{1ex}\textbf{5.50} 
&\textbf{100.00}
&{79.24}
&{2.76}
&\textbf{83.12}	&\textbf{55.65}	&\hspace{1ex}\textbf{6.09}
\\
\hline
\end{tabular}
}
\egroup
\caption{\label{tab:tgt_single} 
  System performance in the single-domain setting on 2000 dialogues in the target domains. 
  }
\end{table}

\begin{table}[t]
\tabcolsep=0.11cm
\small
\bgroup
\def\arraystretch{1.0}
\centerline{
\begin{tabular}{l|ccc|ccc|ccc}
\hline
\multirow{2}*{\textbf{Systems}} & \multicolumn{3}{c}{\texttt{Hotel}} & \multicolumn{3}{c}{\texttt{Train}} &
\multicolumn{3}{c}{Average}\\
~ & 
\textbf{Success} & \textbf{Reward} & \textbf{Turns} & 
\textbf{Success} & \textbf{Reward} & \textbf{Turns}&
\textbf{Success} & \textbf{Reward} & \textbf{Turns}\\
\hline
\rowcolor[gray]{0.95}\multicolumn{10}{c}{\textbf{Few-Shot Models}}\\
\hline
\textsc{Dqn-1k} 
&\hspace{1ex}0.00	&-50.38	&12.38
&\hspace{1ex}2.15	&-56.02 &20.60
&\hspace{1ex}1.08	&-53.20	&16.49
\\
\textsc{Dtqn-1k}
&\hspace{1ex}2.55&	-47.77&	12.84	&\hspace{1ex}8.90&	-45.96&	18.64
&\hspace{1ex}5.73&	-46.87&	15.74
\\
\hline
\rowcolor[gray]{0.95}\multicolumn{10}{c}{\textbf{Adaptive Models}}\\
\hline
\textsc{VanillaDqn}
&\hspace{1ex}0.05	&-58.93	&20.99
&\hspace{1ex}0.00	&-59.00	&21.00
&\hspace{1ex}0.03	&-58.97	&21.00
\\
\textsc{Dqn} 
&\hspace{1ex}3.90&	-53.74	&20.42
&\hspace{1ex}9.85&	-45.52	&19.34
&\hspace{1ex}6.88&	-49.63	&19.88
\\
\textsc{Dtqn}
&\hspace{1ex}4.15   &-53.32 &20.30
&15.40	&-37.93	&18.41
&\hspace{1ex}9.78	&-45.63	&19.36
\\
\textsc{Meta-Dtqn-Sr}
&\hspace{1ex}1.15	&-57.41	&20.79	
&\hspace{1ex}4.35	&-53.01	&20.23
&\hspace{1ex}2.75	&-55.21	&20.51\\
\textsc{Meta-Dtqn}
&\textbf{11.45}	&\textbf{-43.68}    &\textbf{19.42}	
&\textbf{19.30}	&\textbf{-32.81}	&\textbf{17.97}
&\textbf{15.38}	&\textbf{-38.25}	&\textbf{18.70}
\\
\hline
\end{tabular}
}
\egroup
\caption{\label{tab:tgt_cross} 
  System performance in the composite-domain  setting on 2,000 dialogues in the target domains. 
  We show results in \texttt{Hotel} and \texttt{Train} as \texttt{Police} has only single-domain dialogue goal.
}
\end{table}

%% file: table_src.tex
\begin{table}[t]
\tabcolsep=0.11cm
\small
\bgroup
\def\arraystretch{1.0}
\centerline{
\begin{tabular}{l|ccc|ccc|ccc}
\hline
\multirow{2}*{\textbf{Systems}} & \multicolumn{3}{c}{Single} &\multicolumn{3}{c}{Composite}
&\multicolumn{3}{c}{Average}\\
~ & 
\textbf{Success} & \textbf{Reward} & \textbf{Turns} &
\textbf{Success} & \textbf{Reward} & \textbf{Turns}  &
\textbf{Success} & \textbf{Reward} & \textbf{Turns}\\
\hline
\textsc{Dqn} 
&90.20&	65.98&	4.26
&40.40&	\hspace{0.5ex}-3.00&	13.48
&65.30	&31.49	&8.87
\\
\textsc{Dtqn}
&91.70&	68.13&	3.91
&74.85&	42.92&	\hspace{1ex}8.90
&83.28&	55.53&	6.41\\
\textsc{Meta-Dtqn-Sr}
&83.80	&57.00	&5.57
&33.95	&\hspace{-0.75ex}-11.81	&14.55
&58.88	&22.59	&\hspace{-1ex}10.06\\
\textsc{Meta-Dtqn}&
\textbf{93.00}&	
\textbf{69.84}&	
\textbf{3.76}&
\textbf{80.30} &	\textbf{49.95}& \hspace{1ex}\textbf{8.41}
&\textbf{86.65}&\textbf{59.90}	&\textbf{6.09}
\\
\hline
\end{tabular}
}
\egroup
\caption{\label{tab:src} 
  System performance on 2,000 dialogues in the training domains. 
}
\end{table}

%% file: fig_test_replay.tex
\begin{figure*}[t]
  \centering
  \includegraphics[width=\textwidth]{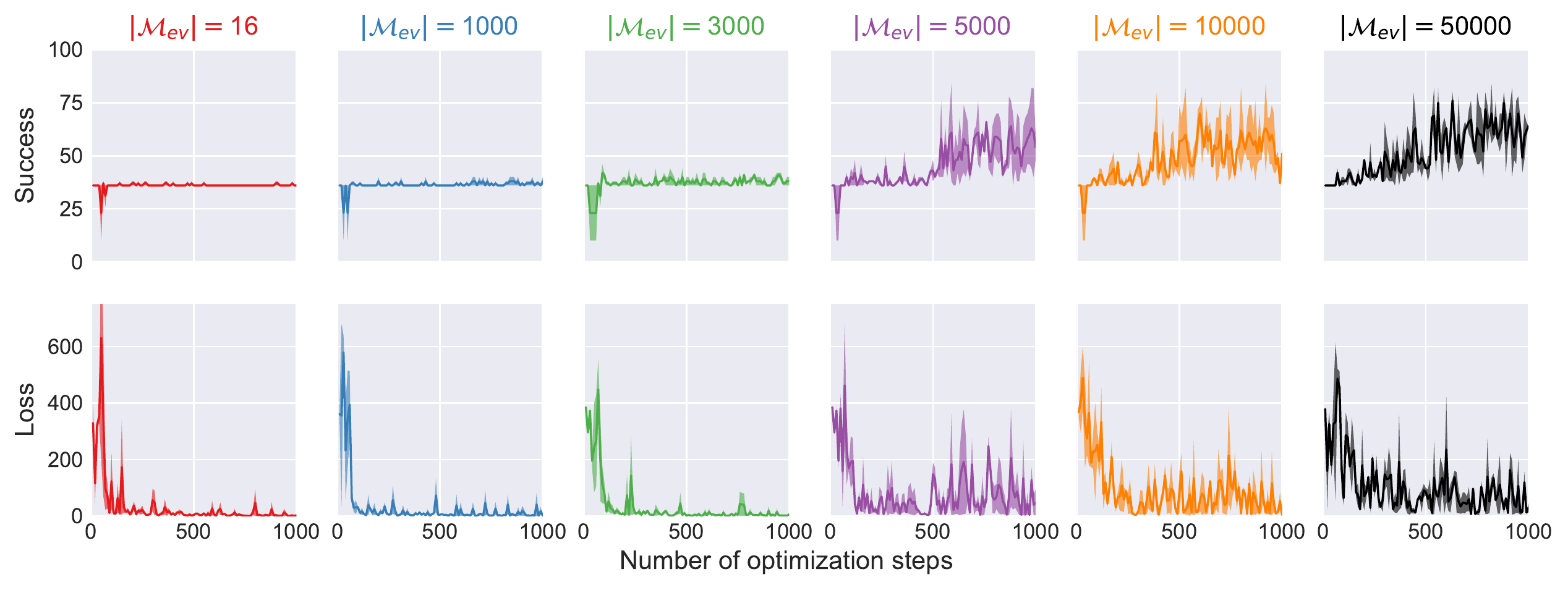}
  \caption{\label{fig:test_mem_size}
  Development success rate (above) and training loss (below) of \textsc{MetaDtqn} with evaluation replay buffer of different sizes on composite-domain tasks. Shadow denotes variance.}
\end{figure*}

%% file: fig_ada_data.tex
\begin{wrapfigure}{L}{0.55\textwidth}
  \centering
  \includegraphics[width=0.55\textwidth]{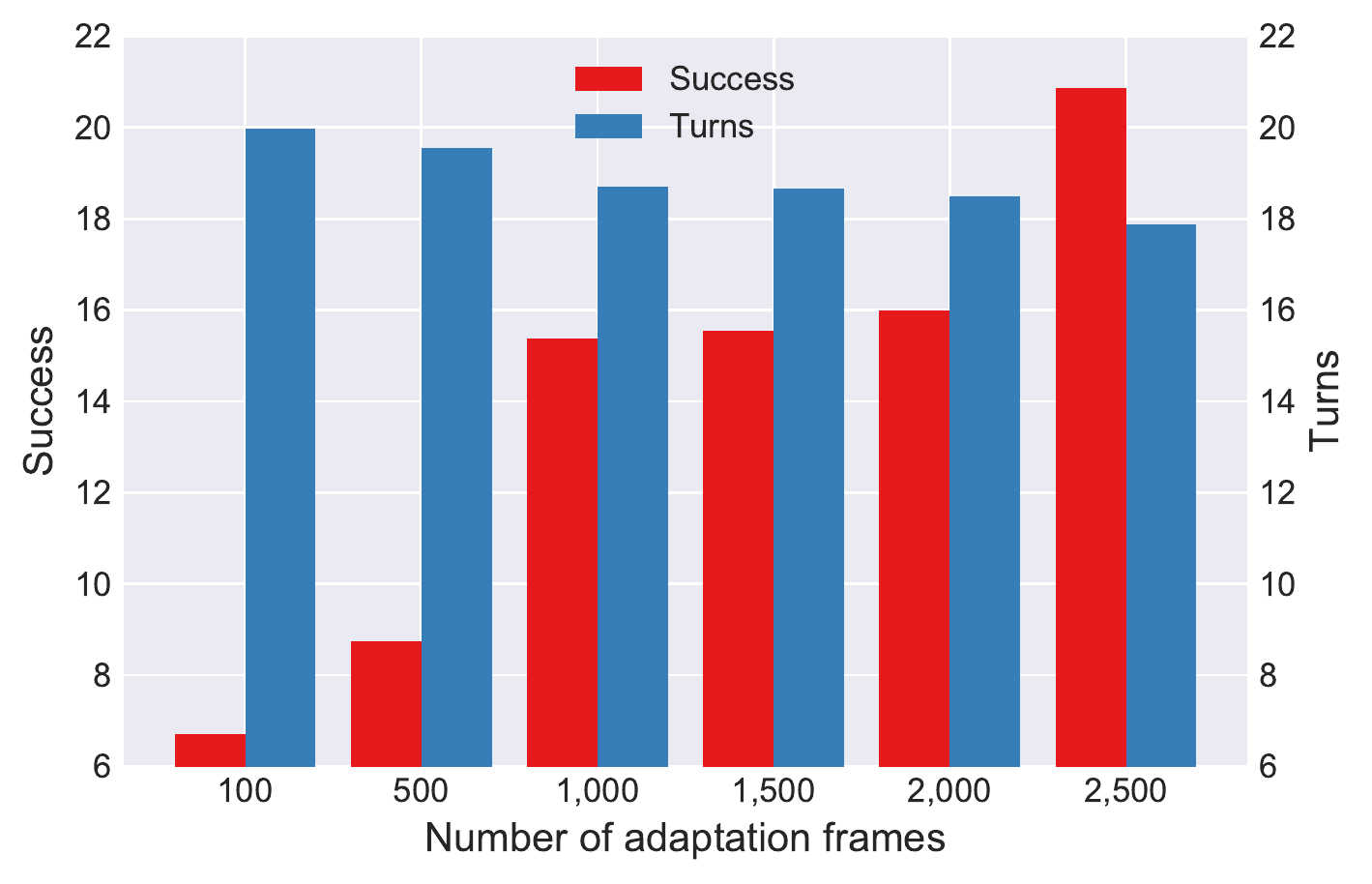}
  \caption{\label{fig:ada_data}
  Performance of \textsc{Meta-Dtqn} with adaptation data of varied sizes on composite tasks.
  }
\end{wrapfigure}